\documentclass[10pt,twocolumn,letterpaper]{article}

\usepackage[pagenumbers]{cvpr} 

\usepackage{graphicx}
\usepackage{amsmath}
\usepackage{amssymb}
\usepackage{booktabs}
\usepackage{multirow}
\usepackage{color}

\PassOptionsToPackage{table,dvipsnames}{xcolor}

\usepackage[table]{xcolor}
\usepackage[dvipsnames]{xcolor}

\usepackage[pagebackref,breaklinks,colorlinks]{hyperref}
\usepackage{comment}

\usepackage[capitalize]{cleveref}
\crefname{section}{Sec.}{Secs.}
\Crefname{section}{Section}{Sections}
\Crefname{table}{Table}{Tables}
\crefname{table}{Tab.}{Tabs.}

\def\cvprPaperID{***} 
\def\confName{CVPR}
\def\confYear{2023}

\definecolor{mygray}{RGB}{230,230,230}

\begin{document}

\title{MarS3D: A Plug-and-Play Motion-Aware Model for Semantic Segmentation on Multi-Scan 3D Point Clouds}

\author{Jiahui Liu$^{1}$\footnotemark[1]~~ Chirui Chang$^{1*}$~~ Jianhui Liu$^1$~~ Xiaoyang Wu$^1$~~ Lan Ma$^2$~~ Xiaojuan Qi$^1$\footnotemark[2]\\
$^1$The University of Hong Kong~~~ $^2$TCL AI Lab\\
\hspace{-12pt}\texttt{\small \{liujh, xjqi\}@eee.hku.hk, crchang@hku.hk, jhliu0212@gmail.com, xywu3@cs.hku.hk, rubyma@tcl.com}
}

\maketitle
\renewcommand{\thefootnote}{\fnsymbol{footnote}}
\footnotetext[1]{Equal contribution.}
\footnotetext[2]{Corresponding author.}

\begin{abstract}

3D semantic segmentation on multi-scan large-scale point clouds plays an important role in autonomous systems. 
Unlike the single-scan-based semantic segmentation task, this task requires distinguishing the motion states of points in addition to their semantic categories. 
However, methods designed for single-scan-based segmentation tasks perform poorly on the multi-scan task due to the lacking of an effective way to integrate temporal information. 
We propose MarS3D, a plug-and-play motion-aware module for semantic segmentation on multi-scan 3D point clouds. 
This module can be flexibly combined with single-scan models to allow them to have multi-scan perception abilities. 
The model encompasses two key designs: the Cross-Frame Feature Embedding module for enriching representation learning and the Motion-Aware Feature Learning module for enhancing motion awareness. 
Extensive experiments show that MarS3D can improve the performance of the baseline model by a large margin.
The code is available at \url{https://github.com/CVMI-Lab/MarS3D}.

\end{abstract}

\section{Introduction}
\label{sec:intro}

3D semantic segmentation on multi-scan large-scale point clouds is a fundamental computer vision task that benefits many downstream problems in autonomous systems, such as decision-making, motion planning, and 3D reconstruction, to name just a few. 
Compared with the single-scan semantic segmentation task, this task requires understanding not only the semantic categories but also the motion states ({\eg}, moving or static) of points based on multi-scan point cloud data.

\begin{figure}[t]
\hspace{-0.2cm}
\includegraphics[width= 0.48\textwidth]{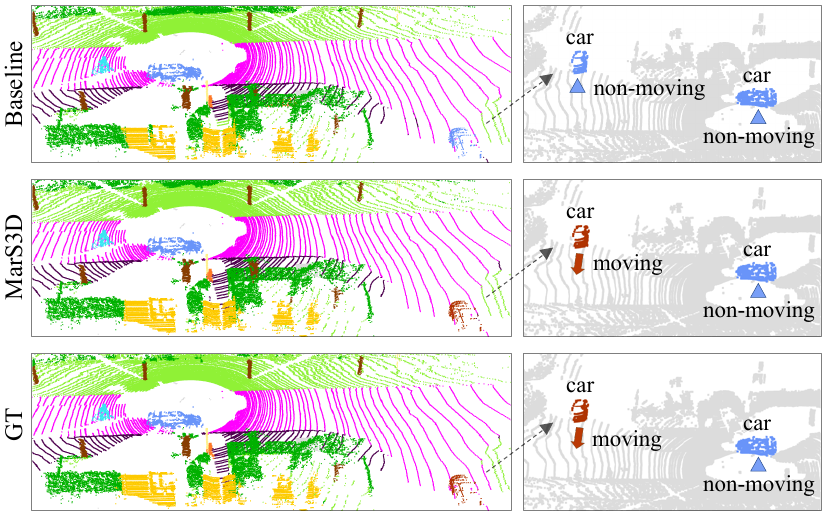}
\vspace{-0.1cm}
\caption{Comparison of our proposed method, MarS3D, with baseline method using SPVCNN~\cite{tang2020searching} as the backbone on SemanticKITTI~\cite{behley2019iccv} dataset. MarS3D achieves excellent results in the classification of semantic categories and motion states, while the baseline method can not distinguish motion well from static.}
\label{intro_fig}
\vspace{-0.2cm}
\end{figure}

In the past few years, extensive research has been conducted on single-scan semantic segmentation with significant research advancements~\cite{hou2022point,xu2021rpvnet,cheng20212,zhou2020cylinder3d,tang2020searching,yan2021sparse,choy20194d}. 
These approaches are also applied to process multi-scan point clouds, wherein multiple point clouds are fused to form a single point cloud before being fed to the network for processing. 
Albeit simple, this strategy may lose temporal information and make distinguishing motion states a challenging problem. 
As a result, they perform poorly in classifying the motion states of objects. As shown in Figure~\ref{intro_fig}, the simple point cloud fusion strategy cannot effectively enable the model to distinguish the motion states of cars even with a state-of-the-art backbone network SPVCNN~\cite{tang2020searching}.
Recently, there have been some early attempts~\cite{schutt2022abstract, shi2020spsequencenet, duerr2020lidar, wang2022meta} to employ attention modules~\cite{shi2020spsequencenet} and recurrent networks~\cite{schutt2022abstract, duerr2020lidar,wang2022meta} to fuse information across different temporal frames. 
However, these approaches do not perform well on the multi-scan task due to the insufficiency of temporal representations and the limited feature extraction ability of the model.

In sum, a systematic investigation of utilizing the rich spatial-temporal information from multiple-point cloud scans is still lacking. 
This requires answering two critical questions: 
(1) how can we leverage the multi-scan information to improve representation learning on point clouds for better semantic understanding? 
and (2) how can the temporal information be effectively extracted and learned for classifying the motion states of objects?

In this paper, we propose a simple plug-and-play \textbf{M}otion-\textbf{a}wa\textbf{r}e \textbf{S}egmentation module for \textbf{3D} multi-scan analysis (\textbf{MarS3D}), which can seamlessly integrate with existing single-scan semantic segmentation models and endow them with the ability to perform accurate multi-scan 3D point cloud semantic segmentation with negligible computational costs. 
Specifically, our method incorporates two core designs: 
First, to enrich representation learning of multi-frame point clouds, we propose a  {Cross-Frame Feature Embedding (CFFE)} module which embeds time-step information into features to facilitate inter-frame fusion and representation learning. 
Second, inspired by the observation that objects primarily move along the horizontal ground plane (\ie, \emph{xy}-plane) in large-scale outdoor scenes, \ie, minimal motion along the \emph{z}-axis, we propose a Motion-Aware Feature Learning (MAFL) module based on Bird's Eye View (BEV), which learns the motion patterns of objects between frames to facilitate effectively discriminating the motion states of objects.

We extensively evaluate our approach upon several mainstream baseline frameworks on SemanticKITTI~\cite{behley2019iccv} and nuScenes~\cite{caesar2020nuscenes} dataset.
It consistently improves the performance of the baseline approaches, \eg, MinkUnet\cite{choy20194d}, by 6.24\% in mIoU on SemanticKITTI with a negligible increase in model parameters, \ie, about 0.2\%. 
The main contributions are summarized as follows:

\begin{itemize}

\vspace{-0.1in}\item We are the first to propose a plug-and-play module
for large-scale multi-scan 3D semantic segmentation, which can be flexibly integrated with mainstream single-scan segmentation models without incurring too much cost.

\vspace{-0.1in}\item We devise a Cross-Frame Feature Embedding module to fuse multiple point clouds while preserving their temporal information,  thereby enriching representation learning for multi-scan point clouds.

\vspace{-0.1in}\item We introduce a BEV-based Motion-Aware Feature Learning module to exploit temporal information and enhance the model's motion awareness, facilitating the prediction of motion states.  

\vspace{-0.1in}\item We conduct extensive experiments and comprehensive analyses of our approach with different backbone models. The proposed model performs favorably compared to the baseline methods while introducing negligible extra parameters and inference time.

\end{itemize}

\section{Related Work}
\label{sec:relat}

\noindent\textbf{Single-scan Outdoor 3D Semantic Segmentation:} 3D single-scan outdoor semantic segmentation is indispensable for autonomous driving. In early work, PointNet~\cite{qi2017pointnet} uses Multi-Layer Perception (MLP) to extract features from input point clouds directly, and PointNet++~\cite{qi2017pointnet++} tries to incorporate multi-scale designs for dense prediction tasks. 
Later, various literature~\cite{qi2017pointnet,qi2017pointnet++,wu2019pointconv, thomas2019kpconv,wang2019dynamic,xu2021paconv} works on designing point-based convolution on either geometric or semantic neighborhoods. 
To handle the large-scale dataset, some works~\cite{graham2017submanifold,choy20194d,zhou2020cylinder3d,tang2020searching,hou2022point,xu2021rpvnet,cheng20212} focus on volumetric features and use 3D convolution to achieve a balance between accuracy and efficiency. 
SparseConv~\cite{graham2017submanifold} and MinkUNet~\cite{choy20194d} are representative works and demonstrate good performance. 
Later, SPVNAS~\cite{tang2020searching} combines voxel and point representations and designs a neural architecture search method to find the optimal model structure. 
Recently, Cylinder3D~\cite{zhou2020cylinder3d} introduces a cylindrical partition to leverage the properties of LiDAR point clouds for enriching the feature information.

The remarkable feature extraction capability enables the above methods to achieve high performance on single-scan tasks. 
To solve the multi-scan task, most of these methods~\cite{zhou2020cylinder3d, hou2022point,xu2021rpvnet,cheng20212,yan20222dpass} first fuse multiple point clouds into one and treat the fused point cloud as a single point cloud for processing. 
Albeit simple, this fusion strategy overlooks important temporal information and entangles moving and non-moving objects, leading to performance degradation.


\begin{figure*}[t]
\hspace{-0.3cm}
\includegraphics[width= 1.01\textwidth]{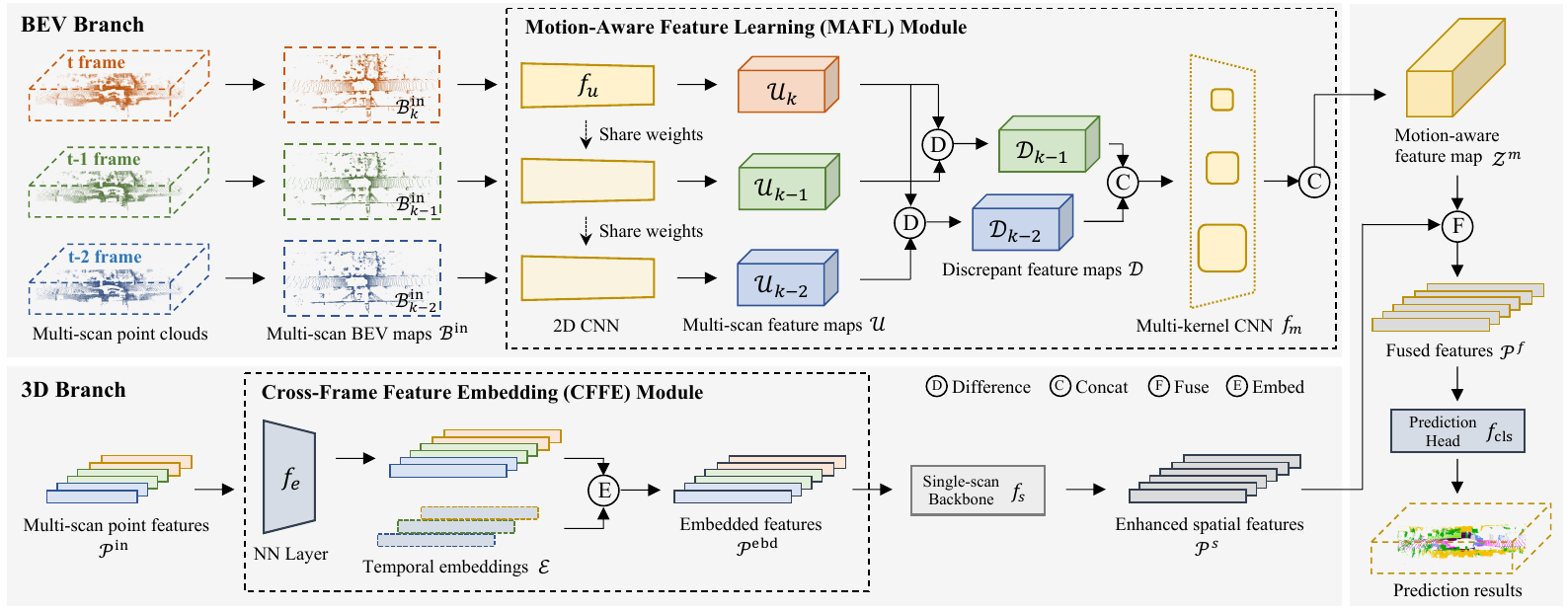}
\vspace{-0.6cm}
\caption{The proposed framework of MarS3D. We take three scan point clouds inputs as illustration. MarS3D contains two branches. One is the BEV branch with 2D BEV representations as input, and it employs a Motion-Aware Feature Learning (MAFL) module to enhance the motion-aware feature learning. The other is the 3D branch that takes multi-scan 3D point clouds as input, enriching the feature representations with our Cross-Frame Feature Embedding (CFFE) module. The fused features of the above two modules are fed into the prediction head and make final predictions on semantic categories and motion states.}
\label{framework}
\vspace{-0.1cm}
\end{figure*}


\vspace{0.1in}\noindent\textbf{Multi-scan Outdoor 3D Semantic Segmentation:}
Compared to single-scan semantic segmentation, the multi-scan task needs to discriminate the moving and stationary states of the objects based on temporal information. 
In addition to the simple fusion strategy discussed above, another stream of approaches ~\cite{schutt2022abstract, shi2020spsequencenet, duerr2020lidar,wang2022meta} attempts to process each point cloud in a sequence separately and fuse the feature representations for temporal modeling. 
For instance, SpSequenceNet~\cite{shi2020spsequencenet} proposes a U-Net-based architecture to extract per-frame features and combine features of two consecutive frames to gather temporal information. 
The fused feature is further fed into the prediction head to produce results. Duerr \etal ~\cite{duerr2020lidar} design a recurrent architecture with a temporal memory alignment module for sequential processing of multiple point clouds.  
TemporalLatticeNet~\cite{schutt2022abstract} proposes to match similar feature patterns between adjacent frames and fuse them temporally.  
However, these approaches cannot fully leverage multiple point clouds to enrich temporal representation learning as the feature extraction is still conducted on each frame separately.

\vspace{0.1in}\noindent\textbf{BEV-based 3D Point Cloud Perception:} Recently, BEV-based representation~\cite{lang2019pointpillars,liu2022BEVfusion,zou2021efficient,zhou2018voxelnet,hendy2020fishing,zhang2020polarnet} has emerged as an effective way to process 3D point cloud due to its efficiency and ease of deployment using 2D convolution operations. 
By projecting a 3D point cloud into the bird's eye view (BEV), BEV-based representation converts 3D representations into 2D to avoid heavy processing in 3D and improve computation and memory efficiency. 
Some representative works include PointPillars~\cite{lang2019pointpillars}  for 3D object detection, BEVfusion~\cite{liu2022BEVfusion} for multi-sensor fusion, BEV projection~\cite{zou2021efficient} for 3D segmentation. 
The above work demonstrates the potential of BEV representations in outdoor 3D scene analysis. Here, we explore utilizing BEV-based representation to extract temporal information for analyzing the motion states of objects.

\vspace{0.1in}\noindent\textbf{2D Video Semantic Segmentation:} In video segmentation, many efforts~\cite{oh2019video,hu2020temporally,robinson2020learning,ding2020every,miao2020memory,bhat2020learning,liu2020efficient} have been made to conduct temporal feature extraction.
Oh \etal~\cite{oh2019video} introduce a space-time memory network to integrate features from adjacent frames for video object segmentation and attains significant performance gains.  
TDNet~\cite{hu2020temporally} exploits temporal redundancy for fast video segmentation by developing an attention propagation method to propagate features to adjacent frames. 
Although they achieve promising results on video segmentation, we experimentally demonstrate these designs are not optimal for multi-scan point cloud segmentation. 
We focus on designing new methods tailored to point clouds for better performance and efficiency.


\section{Problem Statement}

Given a sequence of LiDAR point clouds as inputs, the multi-scan 3D point cloud semantic segmentation task aims to assign a semantic category to each point and predict their motion states (\ie, moving or static). Specifically, a LiDAR point cloud frame contains a set of unordered points that are annotated with labels for training. We denote a pair of training data as  $(\mathcal{P}_{i},\mathcal{L}_{i})=\left\{p_{j},l_{j}\right\}_{j=1}^{N}$ with $p_{j} \in \mathbb{R}^{D_\text{in}}$, where $N$ denotes the number of points. Each point $p_{j}$ contains input descriptors with $D_\text{in}$ dimensions, including point coordinates $(x,y,z)$ and other features such as intensities ($\text{bm}$). The corresponding label $l_{j}$ incorporates both semantic categories and motion states of $p_{j}$. Therefore, points belonging to the same semantic category but possessing distinct motion states are allocated distinct labels.
For a sequence of point clouds $\left\{\mathcal{P}_{i},\mathcal{L}_{i}\right\}_{i=1}^{M}$ that contains $M$ frames, all frames are scanned sequentially in time order, and scanning poses and timestamps are used to align different frames into the same world-coordinate system. In the following, we omit the alignment process for simplicity, where the point clouds in a sequence are calibrated to the same coordinate system by default.

\section{Method}
\label{sec:method}

\subsection{Overview}

An overview of our motion-aware model for multi-scan semantic segmentation, namely MarS3D, is shown in Figure~\ref{framework}.  MarS3D contains a 3D branch for multi-scan spatial representation learning and a BEV branch for motion-aware feature learning. First, the BEV branch takes as inputs $k$ BEV representations $\mathcal{B}^\text{in}=\left\{\mathcal{B}_{i}^\text{in}\right\}_{i=1}^{k}$ that are derived by point cloud polarization (see Section~\ref{sec:bev_branch}) and outputs motion-aware feature map $\mathcal{Z}^m$. The core is a Motion-Aware Feature Learning (MAFL) module (see Section~\ref{sec:bev_branch}) that extracts and leverages BEV features through a dedicated design to produce a motion-aware feature map $\mathcal{Z}^{m}$. Second, the 3D branch takes as inputs the fused $k$ point clouds $\mathcal{P}^{\text{in}} = \left\{\mathcal{P}_{i}^{\text{in}}\right\}_{i=1}^{k}$ and outputs enriched 3D enhanced spatial features $\mathcal{P}^s$. The 3D branch incorporates a cross-frame feature embedding (CFFE) module (see Section~\ref{sec:3d_branch}) to inject temporal information. It outputs embedded features, denoted as $\mathcal{P}^{\text{ebd}} = \left\{\mathcal{P}_{i}^{\text{ebd}}\right\}_{i=1}^{k}$. These features are further processed by a single-scan backbone $f_s$  to yield multi-scan enhanced 3D spatial representations $\mathcal{P}^s$ as:

\begin{equation}
\mathcal{P}^{s}=f_{s}(\mathcal{P}^{\text{ebd}}),
\label{eq:backbone}
\end{equation}
where $\mathcal{P}^{s}\in \mathbb{R}^{D_{z}}$ and $D_{z}$ is the dimension of the output feature of the single-scan backbone.

Finally, the motion-aware feature map $\mathcal{Z}^m$ and enhanced spatial features $\mathcal{P}^s$ are combined to produce the fused features $\mathcal{P}^f$ by aligning the coordinates of the features with the pixels of the motion-aware feature map. The fused features are then fed to the prediction head to produce the final outputs (see Section~\ref{sec:pred}).

\subsection{BEV Branch}
\label{sec:bev_branch}

In the following, we elaborate on the BEV branch, which targets extracting motion-friendly features for motion prediction (see Figure~\ref{framework}). Before delving into the details, we first introduce BEV mapping, which maps a 3D point cloud into a 2D BEV image. Then, we introduce our key Motion-Aware Feature Learning (MAFL) module for motion extraction.

\vspace{0.1in}\noindent\textbf{BEV Mapping:} Each point cloud $\mathcal{P}$ for the BEV branch is pillarized into a three-channel BEV map $\mathcal{B}$ of size $H\times W$. Inspired by~\cite{lang2019pointpillars,liu2022BEVfusion}, for a pillar located at $(x^*,y^*)$ with a pixel grid size of $l_B$, its three-channel feature $b_{(x^*,y^*)}$ consists of the average translation along $x$ and $y$ axis regarding the grid center, and total intensity features (taking SemanticKITTI as an example) among all points inside the pillar, which is formulated as:

\begin{equation}
b_{(x^*,y^*)}=\left[ \frac{1}{N^{*}}\sum_{i}^{N^{*}}\frac{2 \Delta x_i}{l_B}, \frac{1}{N^{*}}\sum_{i}^{N^{*}}\frac{2 \Delta y_i}{l_B}, \sum_{i}^{N^{*}}\text{bm}_{i} \right],
\label{eq:bev-mapping}
\end{equation}
where $N^{*}$ denotes the number of points in current pillar, while $\Delta x_i$ and $\Delta y_i$ denote the translation of the $i$th point along \emph{x-axis} and \emph{y-axis} respectively.

\vspace{0.1in}\noindent\textbf{Motion-Aware Feature Learning Module:} Our Motion-Aware Feature Learning (MAFL) module is illustrated in the BEV branch in Figure~\ref{framework} which is designed to extract motion-aware features for better distinguishing moving/static objects. The input to this module consists of $k$ multi-scan BEV representations $\mathcal{B}^\text{in}=\left\{\mathcal{B}_{i}^\text{in}\right\}_{i=1}^{k}$, where $\mathcal{B}_{k}^\text{in}$ represents the target frame. A lightweight 2D CNN $f_{u}$ with a UNet-like architecture~\cite{ronneberger2015u} is used to extract features $\mathcal{U} = \left\{\mathcal{U}_{i}\right\}_{i=1}^{k}$ from multi-scan inputs as:

\begin{equation}
\left\{\mathcal{U}_{i}\right\}_{i=1}^{k}=\left\{f_{u}(\mathcal{B}_{i}^\text{in})\right\}_{i=1}^{k}.
\end{equation}
Furthermore, to identify moving objects, we take  the difference between the target frame $k$ and the remaining $k-1$ reference frames which  outputs $\mathcal{D} = \left\{\mathcal{D}_{i}\right\}_{i=1}^{k-1}$ as:

\begin{equation}
\left\{\mathcal{D}_{i}\right\}_{i=1}^{k-1}=\left\{\mathcal{U}_{k} - \mathcal{U}_{i}\right\}_{i=1}^{k-1}.
\end{equation}
By doing so, the static objects can be erased, and the dynamic objects are highlighted with a large feature magnitude. Then, $\left\{\mathcal{D}_{i}\right\}_{i=1}^{k-1}$ are channel-wise concatenated to form a new 2D map. Note that objects may have different moving patterns and velocities. Therefore, we design a multi-kernel convolutional network $f_m$ with multiple branches of various kernel sizes to capture objects with various movement patterns. Finally, the outputs from $f_m$ are concatenated to output a motion-aware feature map $\mathcal{Z}^m$.


\begin{table*}[t]
\hspace{-0.3cm}
  \centering \scalebox{0.61}{
  \renewcommand\arraystretch{1.8}
  \begin{tabular} {c|c|c|cc|p{.45cm}p{.45cm}p{.45cm}p{.45cm}p{.45cm}p{.45cm}p{.45cm}p{.45cm}p{.45cm}p{.45cm}p{.45cm}p{.45cm}p{.45cm}p{.45cm}p{.45cm}p{.45cm}p{.45cm}p{.45cm}p{.45cm}p{.45cm}p{.45cm}p{.45cm}p{.45cm}p{.45cm}p{.45cm}}
  \bottomrule[1.5pt]
  \textbf{B.} & \textbf{Method} & \textbf{mIoU} & \textbf{\#param} & \textbf{latency} & $car$ & $bic.$ & $mot.$ & $tru.$ & $ove.$ & $per.$ & $bil.$ & $mol.$ & $roa.$ & $par.$ & $sid.$ & $ogr.$ & $bui.$ & $fen.$ & $veg.$ & $trn.$ & $ter.$ & $pol.$ & $tra.$ & $mca.$ & $mbi.$ & $mpe.$ & $mmo.$ & $mov.$ & $mtr.$ \\ \hline
  \multirow{3}*{\rotatebox{90}{SPVCNN~\cite{tang2020searching}}} & {\em Baseline} & 49.70 & 21.8M & 206ms & 93.9 & 34.4 & 64.7 & 68.0 & 33.0 & 19.7 & 0.0 & 0.0 & 93.6 & 45.2 & 80.1 & 0.2 & 90.3 & 59.7 & 88.4 & 63.5 & 75.6 & 64.1 & 51.9 & 74.3 & 86.7 & 55.0 & 0.0 & 0.0 & 0.0  \\
  ~ & {\em Ours} & 54.66 & 21.9M & 225ms & 95.6 & 52.7 & 77.8 & 79.4 & 51.5 & 27.9 & 0.0 & 0.0 & 94.2 & 51.3 & 82.0 & 0.1 & 91.6 & 65.1 & 89.0 & 68.4 & 76.2 & 65.2 & 51.5 & 80.6 & 94.9 & 68.0 & 0.0 & 3.6 & 0.0  \\
  ~ & \cellcolor{mygray} $\Delta$ & \cellcolor{mygray} \hspace{-3pt}\textcolor[RGB]{5,80,8}{+4.96} & \cellcolor{mygray} \hspace{-3pt}+0.1M & \cellcolor{mygray} \hspace{-3pt}+19ms & \cellcolor{mygray} 1.7 & \cellcolor{mygray} 18.3 & \cellcolor{mygray} 13.1 & \cellcolor{mygray} 11.4 & \cellcolor{mygray} 18.5 & \cellcolor{mygray} 8.2 & \cellcolor{mygray} 0.0 & \cellcolor{mygray} 0.0 & \cellcolor{mygray} 0.6 & \cellcolor{mygray} 6.1 & \cellcolor{mygray} 1.9 & \cellcolor{mygray} \textcolor[rgb]{0.161,0.204,0.710}{0.1} & \cellcolor{mygray} 1.3 & \cellcolor{mygray} 5.4 & \cellcolor{mygray} 0.6 & \cellcolor{mygray} 4.9 & \cellcolor{mygray} 0.6 & \cellcolor{mygray} 1.1 & \cellcolor{mygray} \textcolor[rgb]{0.161,0.204,0.710}{0.4} & \cellcolor{mygray} 6.3 & \cellcolor{mygray} 8.2 & \cellcolor{mygray} 13.0 & \cellcolor{mygray} 0.0 & \cellcolor{mygray} 3.6 & \cellcolor{mygray} 0.0  \\ \hline
  \multirow{3}*{\rotatebox{90}{SparseConv~\cite{graham20183d}}} & {\em Baseline} & 48.99 & 39.2M & 239ms & 94.7 & 24.1 & 54.1 & 69.6 & 43.4 & 17.3 & 0.2 & 0.0 & 93.2 & 45.1 & 79.8 & 0.2 & 89.5 & 61.7 & 87.7 & 62.9 & 74.6 & 63.8 & 50.0 & 73.9 & 85.4 & 53.6 & 0.0 & 0.0 & 0.0   \\
  ~ & {\em Ours} & 54.64 & 39.3M & 253ms & 96.6 & 35.2 & 69.0 & 83.3 & 64.8 & 26.9 & 0.0 & 0.0 & 94.0 & 61.2 & 82.5 & 0.1 & 90.9 & 65.8 & 88.1 & 67.8 & 75.2 & 66.2 & 51.4 & 83.5 & 94.4 & 68.9 & 0.0 & 0.0 & 0.0  \\
  ~ & \cellcolor{mygray} $\Delta$ & \cellcolor{mygray} \hspace{-3pt}\textcolor[RGB]{5,80,8}{+5.65} & \cellcolor{mygray} \hspace{-3pt}+0.1M & \cellcolor{mygray} \hspace{-3pt}+14ms & \cellcolor{mygray} 1.9 & \cellcolor{mygray} 11.1 & \cellcolor{mygray} 14.9 & \cellcolor{mygray} 13.7 & \cellcolor{mygray} 21.4 & \cellcolor{mygray} 9.6 & \cellcolor{mygray} \textcolor[rgb]{0.161,0.204,0.710}{0.2} & \cellcolor{mygray} 0.0 & \cellcolor{mygray} 0.8 & \cellcolor{mygray} 16.1 & \cellcolor{mygray} 2.7 & \cellcolor{mygray} \textcolor[rgb]{0.161,0.204,0.710}{0.1} & \cellcolor{mygray} 1.4 & \cellcolor{mygray} 4.1 & \cellcolor{mygray} 0.4 & \cellcolor{mygray} 4.9 & \cellcolor{mygray} 0.6 & \cellcolor{mygray} 2.4 & \cellcolor{mygray} 1.4 & \cellcolor{mygray} 9.6 & \cellcolor{mygray} 9.0 & \cellcolor{mygray} 15.3 & \cellcolor{mygray} 0.0 & \cellcolor{mygray} 0.0 & \cellcolor{mygray} 0.0  \\ \hline
  \multirow{3}*{\rotatebox{90}{MinkUNet~\cite{choy20194d}}} & {\em Baseline} & 48.47 & 37.9M & 295ms & 93.8 & 23.7 & 48.9 & 90.3 & 41.3 & 18.0 & 0.0 & 0.0 & 92.2 & 32.2 & 78.4 & 0.0 & 89.8 & 55.5 & 88.8 & 63.7 & 77.0 & 63.6 & 50.0 & 69.2 & 83.1 & 52.5 & 0.0 & 0.0 & 0.0  \\
  ~ & {\em Ours} &54.71 & 38.0M & 323ms & 96.4 & 28.4 & 70.0 & 93.9 & 62.7 & 31.6 & 0.0 & 0.0 & 93.8 & 58.5 & 81.7 & 0.1 & 92.6 & 67.6 & 89.0 & 66.7 & 76.4 & 66.5 & 51.6 & 82.6 & 93.1 & 64.4 & 0.0 & 0.1 & 0.0 \\
  ~ & \cellcolor{mygray} $\Delta$ & \cellcolor{mygray} \hspace{-3pt}\textcolor[RGB]{5,80,8}{+6.24} & \cellcolor{mygray} \hspace{-3pt}+0.1M & \cellcolor{mygray} \hspace{-3pt}+28ms & \cellcolor{mygray} 2.6 & \cellcolor{mygray} 4.7 & \cellcolor{mygray} 21.1 & \cellcolor{mygray} 3.6 & \cellcolor{mygray} 21.4 & \cellcolor{mygray} 13.6 & \cellcolor{mygray} 0.0 & \cellcolor{mygray} 0.0 & \cellcolor{mygray} 1.6 & \cellcolor{mygray} 26.3 & \cellcolor{mygray} 3.3 & \cellcolor{mygray} 0.1 & \cellcolor{mygray} 2.8 & \cellcolor{mygray} 12.1 & \cellcolor{mygray} 0.2 & \cellcolor{mygray} 3.0 & \cellcolor{mygray} \textcolor[rgb]{0.161,0.204,0.710}{0.6} & \cellcolor{mygray} 2.9 & \cellcolor{mygray} 1.6 & \cellcolor{mygray} 13.4 & \cellcolor{mygray} 10.0 & \cellcolor{mygray} 11.9 & \cellcolor{mygray} 0.0 & \cellcolor{mygray} 0.1 & \cellcolor{mygray} 0.0  \\ \toprule[1.5pt]
  \end{tabular}
  }
  \vspace{-0.2cm}
  \caption{Quantitative results of the proposed method, MarS3D, on SemanticKITTI~\cite{behley2019iccv} multi-scan public validation set. Combined with different mainstream single-scan 3D point cloud semantic segmentation backbones, MarS3D has a large performance improvement over the corresponding baseline methods without introducing excessive parameters and each-frame inference time. (\textbf{B.} indicates Backbone, full names of the categories are in the supplementary material, {\colorbox{mygray}{\textcolor[rgb]{0.161,0.204,0.710}{blue}}} indicates degradation.)}
  \label{tab:main_result}
\vspace{-0.1cm}
\end{table*}

\subsection{3D Branch}
\label{sec:3d_branch}

The 3D branch uses temporal information to enhance spatial representation learning on 3D point clouds. The core component is the Cross-Frame Feature Embedding (CFFE) module, whose output is fed into the single-scan backbone network to produce enhanced spatial features $\mathcal{P}^s$ (see Figure~\ref{framework}). In the following, we elaborate on the CFFE module to improve spatial representation learning on multi-scan point clouds.

\vspace{0.1in}\noindent\textbf{Cross-Frame Feature Embedding Module:} When the multi-frame point clouds are fused as discussed in the previous sections, points from different time steps are mixed, making it challenging for the final recognition. Inspired by positional embedding~\cite{vaswani2017attention}, we propose a Cross-Frame Feature Embedding (CFFE) module to generate a time-aware embedding and produce consistent features for each point across different timestamps. 
Given $k$ point cloud frames,  $\left\{\mathcal{P}_{i}^\text{in}\right\}_{i=1}^{k}$, we design an embedding neural network layer $f_e$ that maps point clouds into intermediate latent features and $k$ learnable temporal embeddings $\mathcal{E}=\left\{e_{i}\right\}_{i=1}^{k}$ corresponding to $k$ frames respectively. The dimension of $e_{i}$ is the same as the output dimension of the $f_e$. The point-level embedded features $\mathcal{P}^{\text{ebd}}$ are obtained by:

\begin{equation}
\left\{\mathcal{P}_{i}^\text{{ebd}}\right\}_{i=1}^{k}=\left\{e_{i} + f_e(\mathcal{P}_{i}^\text{in})\right\}_{i=1}^{k},
\end{equation}
where element-wise summation is conducted on $e_{i}$ and each point of the corresponding $f_e(\mathcal{P}_{i}^\text{in})$. The obtained features with different temporal embeddings are represented in a point cloud $\{\mathcal{P}_i^\text{ebd}\}_{i=1}^k$ and subsequently fed into the single-scan backbone network $f_s$, following Eq.~\eqref{eq:backbone}, to produce a set of enhanced spatial features $\mathcal{P}^s$.

\subsection{Feature Fusion and Prediction}
\label{sec:pred}

As illustrated in Figure~\ref{framework}, equipped with the motion-aware feature map ($\mathcal{Z}^m$) and enhanced spatial features ($\mathcal{P}^s$), the next step is to fully integrate the representation information from both branches and make predictions.

\vspace{0.1in}\noindent\textbf{Feature Fusion:} The motion-aware feature map $\mathcal{Z}^m \in \mathbb{R}^{H \times W \times D_{z}}$ ($H$: height; $W$: width; $D_z$: number of channels) is obtained from the BEV branch, while the enhanced spatial features $\mathcal{P}^s \in \mathbb{R}^{N \times D_{p}}$ ($N$: number of points; $D_p$: number of channels) are the outputs of the 3D branch. The feature fusion module aggregates information from the above two representations to make subsequent predictions. For each point feature in $\mathcal{P}^s$ with $D_{p}$ dimension, a corresponding pixel from $\mathcal{Z}^m$ can be queried based on the point's 3D location (\emph{x} and \emph{y} coordinates). This pixel serves as an index to extract a $D_z$-dimensional motion-aware feature along the channel dimension from $\mathcal{Z}^m$. Subsequently, the motion-aware feature is concatenated with the point feature, resulting in the fused features denoted as $\mathcal{P}^f \in \mathbb{R}^{N \times (D_z+D_p)}$.

\vspace{0.1in}\noindent\textbf{Prediction:} Based on the correspondence between 2D and 3D, $\mathcal{P}^f$ is then fed into an MLP classifier $f_\text{{cls}}$ to obtain the output $s_\text{{pred}}$:

\begin{equation}
s_\text{pred}=f_\text{{cls}}(\mathcal{P}^f),
\label{eq:cls}
\end{equation}
where $s_\text{pred}$ is the logits for the prediction result of the semantic segmentation task of the input point cloud.


\begin{figure*}[t]
\hspace{-0.3cm}
\includegraphics[width= 1.01\textwidth]{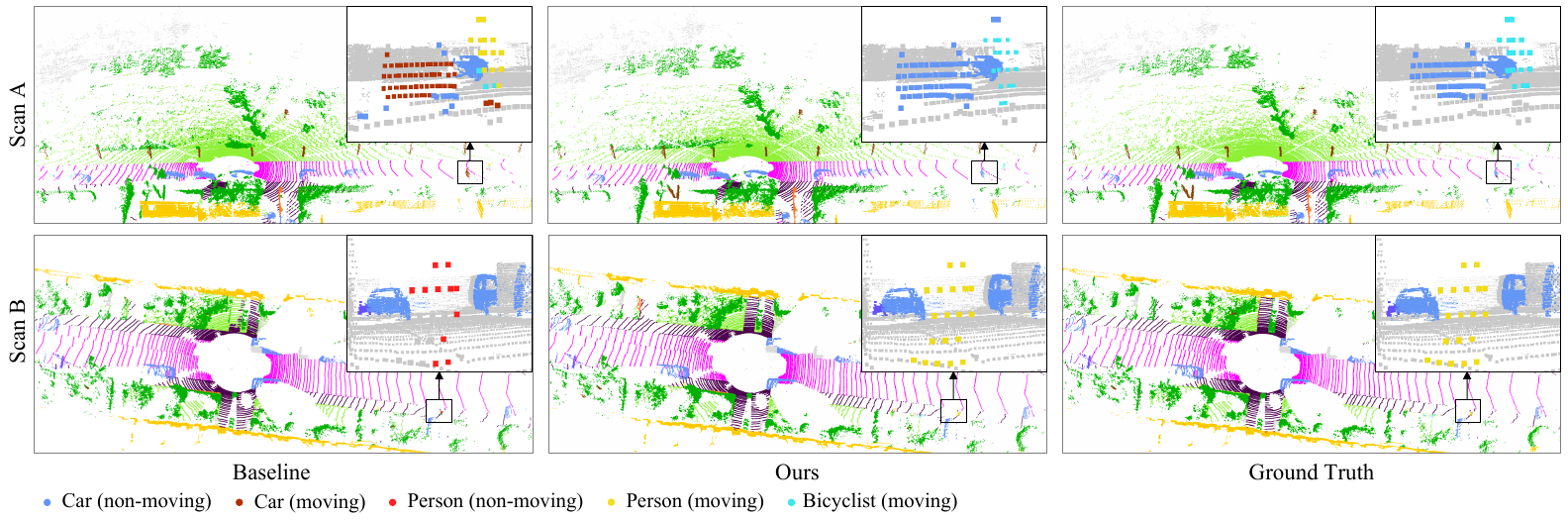}
\vspace{-0.6cm}
\caption{Qualitative results on SemanticKITTI~\cite{behley2019iccv} public validation dataset. With SPVCNN~\cite{tang2020searching} as the backbone, the segmentation results on the SemanticKITTI multi-scan task of the baseline model and our model are shown together with ground truth. At the same time, a specific area containing moving points is magnified and displayed at the top right of each sub-figure.}
\label{main_result}
\vspace{-0.1cm}
\end{figure*}

\begin{figure}[t]
\hspace{-0.3cm}
\vspace{-0.2cm}
\includegraphics[width= 0.49\textwidth]{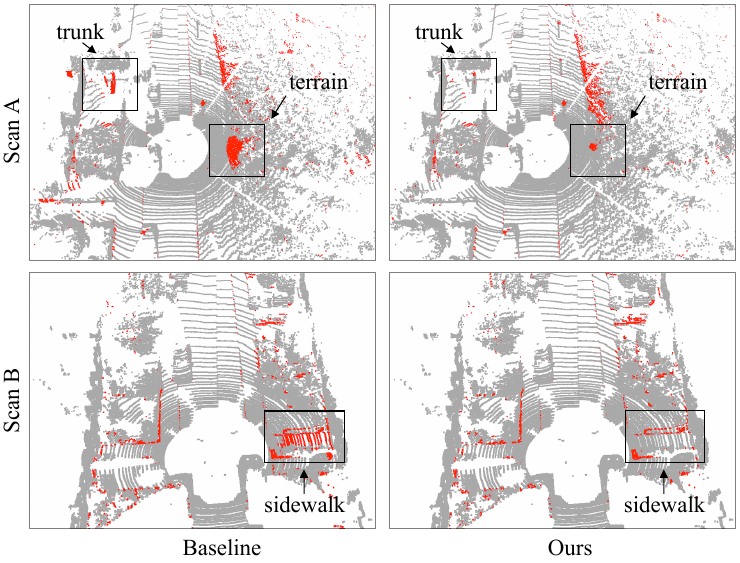}
\caption{Evaluation errors (shown in red) by the baseline methods and MarS3D (using SPVCNN~\cite{tang2020searching} as the backbone) on SemanticKITTI~\cite{behley2019iccv} public validation dataset. MarS3D significantly outperforms the baseline. Contrasting regions in the evaluation errors are highlighted with boxes and corresponding categories.}
\label{error_map}
\vspace{-0.1cm}
\end{figure}

\subsection{Model Training and Inference}

During training, the obtained fused features $\mathcal{P}^f$ are fed into two classification heads: the category-aware classification head and the motion-aware classification head (a binary classifier), which outputs the predicted logits $s_\text{{pred}}^{c}$ and $s_\text{{pred}}^{m}$ (more details are included in the supplementary material). First, with the ground truth labels of semantic categories $\mathcal{L}_\text{{GT}}^{c}$, back-propagation is performed for parameters optimization using Cross Entropy (CE) loss $L_c$ for semantic categories classification:

\begin{equation}
L_c = \text{CE}(s_\text{pred}^c,\mathcal{L}_\text{{GT}}^{c}), 
\end{equation}
where $\text{CE}(\cdot,\cdot)$ is the cross-entropy loss, and $s_\text{pred}^c$ is the prediction probability for each point. Then, the Binary Cross Entropy (BCE) loss $L_m$ is used for motion states classification with motion states ground truth $\mathcal{L}_\text{{GT}}^{m}$:

\begin{equation}
L_m = \text{BCE}(s_\text{pred}^m,\mathcal{L}_\text{GT}^{m}), 
\end{equation}
where $\text{BCE}(\cdot,\cdot)$ indicates the binary cross-entropy loss. 

The final objective function $L$ of the optimization is:

\begin{equation}
L = \omega_c \cdot L_c + \omega_m \cdot L_m,
\end{equation}
where $\omega_c$ and $\omega_m$ are the weights of the two losses ($L_c$ and $L_m$) respectively.

During inference, the final prediction result is determined using the logits produced by both classification heads. When presented with an input sample, the motion-aware classification head will identify the motion state of the input point only if the category-aware classification head recognizes the point as belonging to a class with the potential to move.

\section{Experiments}
\label{sec:exper}

\noindent
\textbf{Datasets and Evaluation Metric:} We evaluate our method on SemanticKITTI~\cite{behley2019iccv} and nuScenes~\cite{caesar2020nuscenes}. For SemanticKITTI, the multi-scan setting is fully supervised and contains 25-category (6 moving categories and 19 static categories) with high-quality semantic annotations. The annotations are based on the KITTI dataset~\cite{Geiger2012CVPR}. It comprises 22 point cloud sequences. For nuScenes, we propose a new multi-scan setting based on the 'lidar-seg' task (16 semantic categories) without reference frame supervision. We use its object-level velocity to construct a multi-scan segmentation dataset with 24 categories (8 moving and 16 static categories). More details are provided in the supplementary material. To assess the effectiveness of our proposed method and make comparisons with baselines and other methods, we use the mean Intersection over Union (mIoU) as the evaluation metric.

\vspace{0.1in}
\noindent
\textbf{Implementation Details:} 
Our model is designed as a plug-and-play module that provides motion-aware features to enhance the backbone learned features. For the backbone model, we consider SPVCNN~\cite{tang2020searching}\footnote{\url{https://github.com/mit-han-lab/spvnas}}, SparseConv~\cite{graham20183d}\footnote{\url{https://github.com/traveller59/spconv}}, and MinkUNet~\cite{choy20194d}\footnote{\url{https://github.com/NVIDIA/MinkowskiEngine}}. Following previous works~\cite{zhou2020cylinder3d,hou2022point,schutt2022abstract}, we use the current and its previous two frames as input. The size of the BEV representation is set to $501 \times 301$, and the multiple kernel sizes in the MAFL module are set as 1, 3, and 5, respectively. We set the embedded feature dimension in the CFFE module to 18, and the data augmentations are the same as the standard settings. All the models are trained on GeForce RTX 3090 GPUs, and the inference latency is recorded using a single GeForce RTX 3090 GPU.

\begin{table*}[t]
\hspace{-0.2cm}
  \centering \scalebox{0.64}{
  \renewcommand\arraystretch{1.8}
  \begin{tabular}{c|c|p{.45cm}p{.45cm}p{.45cm}p{.45cm}p{.45cm}p{.45cm}p{.45cm}p{.45cm}p{.45cm}p{.45cm}p{.45cm}p{.45cm}p{.45cm}p{.45cm}p{.45cm}p{.45cm}p{.45cm}p{.45cm}p{.45cm}p{.45cm}p{.45cm}p{.45cm}p{.45cm}p{.45cm}p{.45cm}}
    \bottomrule[1.2pt]
    \textbf{Method} & \hspace{0pt} \textbf{mIoU} & $car$ & $bic.$ & $mot.$ & $tru.$ & $ove.$ & $per.$ & $bil.$ & $mol.$ & $roa.$ & $par.$ & $sid.$ & $ogr.$ & $bui.$ & $fen.$ & $veg.$ & $trn.$ & $ter.$ & $pol.$ & $tra.$ & $mca.$ & $mbi.$ & $mpe.$ & $mmo.$ & $mov.$ & $mtr.$ \\
    
    \hline
    SpSequenceNet~\cite{shi2020spsequencenet} & 43.1 & 88.5 & 24.0 & 26.2 & 29.2 & 22.7 & 6.3 & 0.0 & 0.0 & 90.1 & 57.9 & 73.9 & 27.1 & 91.2 & 66.8 & 84.0 & 66.0 & 65.7 & 50.8 & 48.7 & 53.2 & 41.2 & 26.2 & 36.2 & 2.3 & 0.1  \\
    TemporalLidarSeg~\cite{duerr2020lidar} & 47.0 & 92.1 & 47.7 & 40.9 & 39.2 & 35.0 & 14.4 & 0.0 & 0.0 & 91.8 & 59.6 & 75.8 & 23.2 & 89.8 & 63.8 & 82.3 & 62.5 & 64.7 & 52.6 & 60.4 & 68.2 & 42.8 & 40.4 & 12.9 & 12.4 & 2.1  \\
    TemporalLatticeNet~\cite{schutt2022abstract} & 47.1 & 91.6 & 35.4 & 36.1 & 26.9 & 23.0 & 9.4 & 0.0 & 0.0 & 91.5 & 59.3 & 75.3 & 27.5 & 89.6 & 65.3 & 84.6 & 66.7 & 70.4 & 57.2 & 60.4 & 59.7 & 41.7 & 51.0 & 48.8 & 5.9 & 0.0  \\
    Meta-RangeSeg~\cite{wang2022meta} & 49.5 & 90.1 & 52.7 & 43.9 & 30.3 & 35.4 & 14.3 & 0.0 & 0.0 & 90.7 & 63.3 & 74.7 & 26.9 & 90.5 & 63.5 & 83.0 & 67.0 & 67.7 & 56.4 & 64.4 & 64.5 & 56.1 & 55.0 & 24.4 & 20.3 & 3.4  \\
    KPConv~\cite{thomas2019kpconv} & 51.2 & 93.7 & 44.9 & 47.2 & 43.5 & 38.6 & 21.6 & 0.0 & 0.0 & 86.5 & 58.4 & 70.5 & 26.7 & 90.8 & 64.5 & 84.6 & 70.3 & 66.0 & 57.0 & 53.9 & 69.4 & 67.4 & 67.5 & 47.2 & 4.7 & 5.8  \\
    \hline
    {\em Baseline} & 49.2 & 89.8 & 39.4 & 34.0 & 39.4 & 21.0 & 8.9 & 1.8 & 0.0 & 89.1 & 62.0 & 72.4 & 12.9 & 90.5 & 63.9 & 84.6 & 68.4 & 68.7 & 58.9 & 60.1 & 69.3 & 63.5 & 58.7 & 56.5 & 9.5 & 3.6 \\
    {\em Ours} & 52.7 & 95.1 & 49.2 & 49.5 & 39.7 & 36.6 & 16.2 & 1.2 & 0.0 & 89.9 & 66.8 & 74.3 & 26.4 & 92.1 & 68.2 & 86.0 & 72.1 & 70.5 & 62.8 & 64.8 & 78.4 & 67.3 & 58.0 & 36.3 & 10.0 & 5.1  \\
    \toprule[1.2pt]
  \end{tabular}
  }
  \vspace{-0.3cm}
  \caption{Comparison with the state-of-the-art models on SemanticKITTI multi-scan benchmark (official test set). MarS3D (with SPVCNN~\cite{tang2020searching} as the backbone) significantly outperforms these models for multi-scan tasks. (full names of the categories are in the supplementary material.)}
  \label{tab:cop_sota}
  \vspace{-0.1cm}
\end{table*}

\subsection{Main Results}

\noindent\textbf{Comparison with Baseline Methods:} We evaluate the performance of our proposed method on multi-scan benchmarks of SemanticKITTI~\cite{behley2019iccv} and nuScenes~\cite{caesar2020nuscenes}. The baseline method uses the same backbone to process multi-scan point clouds. We then compare this baseline approach to the same backbone augmented with MarS3D for a fair evaluation. As shown in Table~\ref{tab:main_result}, MarS3D significantly improves the performance over baseline methods on the public validation set of SemanticKITTI. With the most lightweight network, SPVCNN~\cite{tang2020searching}, MarS3D brings a 4.96\% improvement while introducing less than 0.5\% additional parameters. Particularly, consistent performance gains are observed on the dynamic object classes with non-moving/moving properties (\ie, car \& moving car, person \& moving person). 
This shows that the proposed BEV branch is both lightweight and powerful. Further, to verify the generalizability of the model, we offer a multi-scan task based on nuScenes 'lidar-seg' dataset~\cite{caesar2020nuscenes}. Our method outperforms the baseline (using MinkUNet~\cite{choy20194d} as the backbone) by a significant margin, achieving 64.83\% mIoU compared to the baseline's 61.90\%. This improvement is achieved with a negligible increase in each-frame inference time from 53ms to 58ms.

\vspace{0.1in}\noindent\textbf{Comparison with State-of-the-Art Methods:} Compared to various models applied to multi-scan tasks, MarS3D (with SPVCNN as the backbone) is evaluated on the SemanticKITTI multi-scan benchmark\footnote{\url{http://www.semantic-kitti.org/tasks} - Semantic Segmentation - Multiple Scans}. As shown in Table~\ref{tab:cop_sota}, the proposed approach has demonstrated superior performance, with a 1.5\% increase in mIoU compared to the current state-of-the-art method, KPConv~\cite{thomas2019kpconv}. Furthermore, our method performs similarly or better than other state-of-the-art models across nearly all categories.

\vspace{0.1in}\noindent\textbf{Qualitative Comparisons:} Quantitative results are shown in Figure~\ref{main_result}. The baseline model suffers from mistaking the status of static cars (Figure~\ref{main_result}: Scan A) and moving persons (Figure~\ref{main_result}: Scan B). It even recognizes the moving bicyclist as a moving person (Figure~\ref{main_result}: Scan A). In contrast, MarS3D can circumvent such category and motion state discrimination errors. In addition to achieving better qualitative results in motion states, MarS3D also outperforms the baseline in semantic categories. We visualize the error maps (errors are shown in red) of the baseline and our method in Figure~\ref{error_map}, where bounding boxes indicating specific categories highlight the differences between our method and the baseline. These improvements indicate that MarS3D has stronger semantic feature extraction capabilities compared to the baseline model, resulting in better segmentation performance even for immobile objects.

\begin{figure}[t]
\hspace{0.1cm}
\includegraphics[width= 0.44\textwidth]{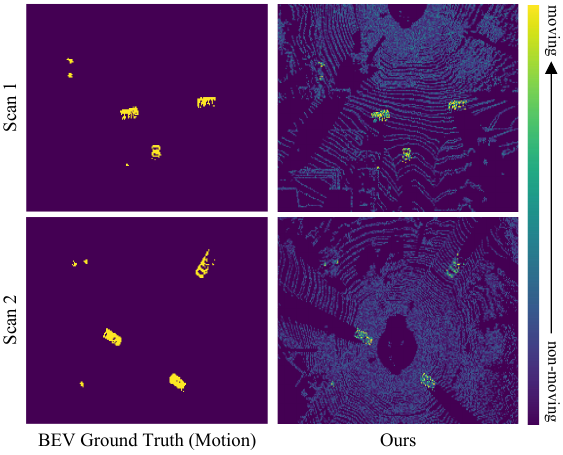}
\vspace{-0.3cm}
\caption{Sampled discrepant feature maps demonstrate that the region that contains moving points are represented higher activation values than other regions.}
\label{BEV_map}
\vspace{-0.2cm}
\end{figure}

\subsection{Ablation Studies}

In this section, we conduct comprehensive ablation experiments on the SemanticKITTI multi-scan validation set to examine the effects of each component in our proposed method. As shown in Table~\ref{tab:ablation}, we gradually add three components to the baseline method, including the CFFE module, vanilla BEV branch (introduce only BEV representations and 2D CNN without the MAFL module), and the MAFL module to illustrate the effectiveness of our designs.

\vspace{0.1in}\noindent\textbf{Effiectiveness of CFFE Module:} As shown in Table~\ref{tab:ablation}, we first employ only the CFFE module with the baseline model and observe a significant performance improvement. For instance, with SPVCNN~\cite{tang2020searching} as the backbone, the introduction of the CFFE module led to a boost in performance by 1.1\% in terms of mIoU. This confirms the efficacy of the CFFE module for multi-frame representation learning.

\vspace{0.1in}\noindent\textbf{Effiectiveness of BEV Representation:} We also observe a significant performance improvement when we add the vanilla BEV branch alone to the baseline model. Using SPVCNN~\cite{tang2020searching} as the backbone, the BEV branch results in a 3.7\% increase in mIoU. This confirms the importance of BEV representations in extracting motion-aware information. Furthermore, when both the CFFE module and the vanilla BEV branch are added to the baseline model, there is a further increase in performance, demonstrating that the two components complement each other.

\vspace{0.1in}\noindent\textbf{Effiectiveness of MAFL Module:} In the final experiment presented in Table~\ref{tab:ablation}, we study the impact of our proposed MAFL module on the BEV representations, resulting in an increase of around 5.0\% from the baseline model (using SPVCNN~\cite{tang2020searching} as the backbone). By stacking all the proposed components together, our final solution (marked with a grey box in Table~\ref{tab:ablation}) reaches its pinnacle in performance. The improvement in other backbones is also significant, and more statistical results are provided in the supplementary material. In addition, the multi-channel discrepant feature map represents the differences between the extracted feature maps of BEV representations from two point clouds. For a discrepant feature map generated during inference, we randomly sample the channels pixel by pixel. The average absolute activation values of pixels with the same motion state in the sampled channels are re-scaled and aggregated into a single-channel feature map as shown in Figure~\ref{BEV_map}. The sampled discrepant feature map has higher activation on pixels containing moving points compared to other regions. This clearly demonstrates that the MAFL module clearly distinguishes between regions that contain moving points and those that do not.


\begin{table}[t]
\hspace{-0.2cm}
    \centering \scalebox{0.97}{
    \renewcommand\arraystretch{1.2}
    \begin{tabular}{c|ccc|c}
        \bottomrule[1pt]
        Method & CFFE & BEV & MAFL & mIoU(\%) \\ 
        \hline
        \emph{baseline} & \hspace{0pt} -  & \hspace{0pt} -  & \hspace{0pt} -  & \hspace{0pt} 49.7 / 49.0 / 48.5 \\
        \hline
        \multirow{4}*{\emph{proposed}} & \hspace{0pt} \checkmark & \hspace{0pt} -  & \hspace{0pt} -  &  \hspace{0pt} 50.8 / 51.3 / 52.1  \\
        ~ & \hspace{0pt} - & \hspace{0pt} \checkmark  & \hspace{0pt} -  & \hspace{0pt} 53.4 / 53.9 / 53.1 \\
        ~ & \hspace{0pt} \checkmark  & \hspace{0pt} \checkmark  & \hspace{0pt} -  & \hspace{0pt} 53.6 / 54.2 / 54.1  \\
        ~ & \cellcolor{mygray} \checkmark  & \cellcolor{mygray}  \checkmark  & \cellcolor{mygray}   \checkmark & \cellcolor{mygray} 54.7 / 54.6 / 54.7 \\ \toprule[1pt]
    \end{tabular}
  }
    \vspace{-0.3cm}
    \caption{Ablation studies on different backbones (SPVCNN / SparseConv / MinkUNet) on SemanticKITTI~\cite{behley2019iccv} public validation dataset. The effectiveness of different designs is demonstrated step-by-step and our method is marked with a gray box.}
    \label{tab:ablation}
\vspace{-0.2cm}
\end{table}

\begin{table}[t]
    \centering \scalebox{0.75}{
    \renewcommand\arraystretch{1.3}
    \begin{tabular}{c|cccc}
        \bottomrule[0.9pt]
        \textbf{Method} & Vanilla BEV & TDNet~\cite{hu2020temporally} & STM~\cite{oh2019video} & \emph{Ours} \\ 
        \hline
        \textbf{mIoU}($\%$) & 53.36 & 53.56 \textcolor[RGB]{5,80,8}{\small(+0.20)} & 53.64 \textcolor[RGB]{5,80,8}{\small(+0.28)} & 54.66 \textcolor[RGB]{5,80,8}{\small(+1.30)} \\
        \toprule[0.9pt]
    \end{tabular}
        }
    \vspace{-0.3cm}
    \caption{Comparing our method (using SPVCNN~\cite{tang2020searching} as the backbone) with different 2D temporal semantic segmentation approaches the public validation dataset of SemanticKITTI~\cite{behley2019iccv}.}
    \label{tab:com_2d}
\vspace{-0.1cm}
\end{table}


\subsection{More Comparison on Temporal Segmentation}

As for BEV representation learning, the proposed MAFL module is utilized on extracted 2D feature maps for capturing inter-frame temporal information of point clouds. We further conduct comparative experiments by replacing the MAFL module with its 2D counterparts, \ie, STM~\cite{oh2019video} and TDNet~\cite{hu2020temporally}. These 2D modules have been widely utilized for preserving temporal information in 2D tasks. The results of the comparative experiments using SPVCNN~\cite{tang2020searching} as the backbone are presented in Table~\ref{tab:com_2d}. According to the evaluation results, it can be concluded that the MAFL module shows superior performance compared to the other models, thereby confirming its remarkable effectiveness. This shows that the proposed MAFL module is better suited for the specific task of handling motion-aware semantic segmentation in 3D point clouds.

\subsection{Limitations and Failure Cases Analysis}

Although MarS3D demonstrates impressive overall performance on the SemanticKITTI multi-scan benchmark, some limitations can be identified from the quantitative results in Table~\ref{tab:main_result} and Table~\ref{tab:cop_sota}. The SemanticKITTI training dataset has an imbalanced distribution of point categories (more details are included in the supplementary material), which causes MarS3D to perform poorly on long-tailed categories due to insufficient training data on these categories. As shown in Table~\ref{tab:cop_sota}, MarS3D fails to show effects on several long-tail categories (\ie, bicyclist, motorcyclist, moving-other-vehicle, and moving-truck). Exploring solutions to address the long-tail problem is a promising research direction for the work. Since our model assumes planar motion, it may not perform well in scenarios where objects move in non-planar ways, such as on steep terrains or on non-planar surfaces.


\section{Conclusion}
\label{sec:conclu}

In this paper, we propose MarS3D, a novel plug-and-play motion-aware model for 3D multi-scan point cloud semantic segmentation. 
The Motion-Aware Feature Learning (MAFL) module, based on BEV representations, is designed to facilitate extracting motion-aware representations. 
Additionally, the Cross-Frame Feature Embedding (CFFE) module is introduced to improve representation learning by embedding time-step information into features, thus preserving rich temporal information.
Extensive experiments and ablation studies demonstrate that MarS3D significantly improves multiple 3D semantic segmentation baselines while introducing minimal overheads. 
In comparison to state-of-the-art methods designed for multi-scan tasks, MarS3D achieves superior performance and offers significant improvements over baseline methods.
The proposed MarS3D model demonstrates the potential for effectively incorporating motion-awareness into 3D point cloud semantic segmentation tasks, providing a strong foundation for further research and development in this area.



\vspace{0.2in}\noindent\textbf{Acknowledgments:} 
This work has been supported by Hong Kong Research Grant Council - Early Career Scheme (Grant No. 27209621), General Research Fund Scheme (Grant no. 17202422), and RGC matching fund scheme (RMGS). Part of the described research work is conducted in the JC STEM Lab of Robotics for Soft Materials funded by The Hong Kong Jockey Club Charities Trust.

{\small
\bibliographystyle{ieee_fullname}
\bibliography{latex/Main}
}

\end{document}